\documentclass{article}

\usepackage[final]{corl_2026} % Uncomment for the camera-ready ``final'' version.

\usepackage{amsmath,amssymb,bbold,graphicx,booktabs,multirow}
\usepackage[table]{xcolor}
\definecolor{crsblue}{RGB}{219,234,250}

\usepackage{adjustbox}

\title{Kinematic Constraint-Referenced Evaluation for Articulated Object Manipulation}

\author{
  Baptiste Bonin$^{1, 2}$ \quad
  Caro Strickland$^{1,2}$ \quad
  Audrey Durand$^{1,2,3}$ \quad
  \\ \\
  $^{1}$Université Laval \qquad
  $^{2}$Mila -- Quebec AI Institute\qquad
  $^{3}$CIFAR AI Chair
}

\begin{document}
\maketitle

\begin{abstract}
Manipulation policies are evaluated predominantly by binary task success, which cannot tell whether the commanded gesture is the one the object's
kinematics admit. We introduce Constraint-Referenced Strategy (CRS), which projects the commanded end-effector twist onto the object's joint axis,
separating progress along the joint from off-constraint waste. CRS requires a joint model and a joint-following scripted controller derived from it, but no recorded demonstration or learned judge. On 74 simulated articulated objects, success registers three transfer failures as lower rates but attributes none, while the two CRS channels give each a distinct signature.
A door-trained behavior-cloning policy opens drawers in 59 of 110 episodes, and in these passes it commands about $40\times$ the off-axis rotation of a
drawer-trained policy's passes ($1.27$ vs.\ $0.032$~rad/s). Relative to a scripted controller scored with the same axis, CRS verdicts are unchanged
under joint-axis errors up to $30^\circ$; absolute thresholds tolerate about $8^\circ$.
\end{abstract}

\keywords{Fine-grained manipulation, Evaluation, Articulated objects}

\section{Introduction}

%\audrey{I have proposed a split of the first sentence and reformulations to stick to a fixed vocabulary (here, prevent jumping between ``binary task success'' and ``standard metrics''.}
% Manipulation policies are evaluated almost universally by binary task success \citep{gu2023maniskill2, geng2022gaparnet, li2024simplerenv}, \audrey{What follows is not a fact: it is what we are arguing. I would split the sentence here and make sure that what follows make it clear that this is what we argue in this paper (not a common known truth).} and that evaluation measures the wrong thing for generalization: whether the task was accomplished, not whether the commanded gesture is right for the target's kinematics. 
Manipulation policies are evaluated predominantly by binary task success \citep{taomaniskill3, li2024simpler}. %\caro{Are these two papers enough to claim that manipulation policies are predominantly evaluated by binary task success?} \baptiste{due to the ref are including into the 4 pages, i tried to limit the ref we do. This is two of the main benchmark use by the community so i assumed it would be enough} \caro{fair! If you wanted to add one more, this might be a good one: ``Robot Learning as an Empirical Science: Best Practices for Policy Evaluation'', but it should be totally fine as-is.}. %We argue that a good evaluation should also measure the commanded gesture against the target's kinematics. %in order to capture what is important for generalization, a good evaluation should instead measure the quality of the commanded gesture given the target's kinematics.
Success aggregates perception, grasping, execution, and strategy into a single bit, so two policies can fail for different reasons and obtain the same score (0). Success can also mislead, reporting a healthy-looking score (1) while the commanded gesture is wrong. For example, a behavior-cloning policy trained on doors opens held-out drawers in more than half of its episodes with the arcing gesture typical of doors, a rotation the drawer's joint does not admit, and success scores each of these episodes as a pass.  A good evaluation should measure the commanded gesture against the target's kinematics. %Standard metrics report this as an unqualified success. %\caro{for understanding, consider adding something like: for example, a policy transferred to an unseen hinged door may pull straight rather than along the hinge's arc, succeeding by exploiting slack in the joint while committing to a gesture that is kinematically impractical. Standard metrics, however, report this as an unqualified success.}

%\audrey{What are diagnostic families? Seems like a sentence is missing to create a link with the previous paragraph.}
%To address these limitations, richer metrics than binary task success have been proposed  \citep{wang2025roboeval, xu2026metafine}. Yet none of them has become the main metric of a manipulation benchmark. \cite{liu2023libero} reports a success rate per suite, while \cite{gu2023maniskill2, li2024simplerenv, james2020rlbench} report success rates.

%Benchmarks evaluate the same policy across dozens of tasks and objects, so the metric must be task-independant.%: it should require no additional specification when a new object or task is introduced, and it should be defined on a scale that can be aggregated across tasks.
%Most of the richer metrics fails this test: demonstration-referenced distances require a recording of the object \cite{li2022dtw, cimse}, graded partial credit needs predefined stages for the mechanism \cite{an2023rgbmanip,vlmsubgoals}, learned evaluators need a task-specific trained judge \cite{duan2025aha}, and human preference needs an annotator for each comparison \cite{atreya2025roboarena}. Execution-quality measures are task-independent, but without a model of the object, they cannot distinguish motion that advances a constraint from motion that fights it. %Binary success is likewise task-independent, but says nothing about the gesture, which is precisely what makes it the headline metric.

Richer evaluation metrics exist, but most require additional information: demonstration-referenced distances need a recording on each object \cite{li2023dtw}, subgoal scoring needs stages defined per task \cite{elmallah2025scorestepsnotjust}, and learned evaluators need a trained judge \cite{duan2025aha}. Across the dozens of tasks and objects a benchmark covers, this cost adds up. MetaFine \cite{xu2026metafine}, %\caro{Nitpicky comment: I feel like it might be a little awkward to say ``the closest to ours is MetaFine'' before formally proposing the contribution; i.e., the reader doesn't know what ``ours'' implies yet?} \audrey{I agree, plus it's unclear from what follows what distinguishes CRS from MetaFine. Instead of saying that it's close to us, I would simply say that it considers the commanded gesture in the evaluation, but... (focus on it's limitations)...}
does evaluate the gesture, not only the outcome: it composes tasks from atomic skills whose checks reference the object's geometry (a rotation passes only about the prescribed axis) and scores behavior by per-stage success and trajectory stability of the commanded actions. Two gaps remain, its geometry checks are pass/fail at a skill-specific tolerance, and its stability score, though task-independent and computed on the command, has no model of the object. Neither returns a graded, signed reading of how much of a command advances the joint and how much fights it. This reading must come from the command: on an articulated part, realized motion is admissible by construction, since a drawer can only slide along its rail, so off-constraint behavior leaves no trace in the part's motion, only in the command and the constraint forces.

\begin{figure}[t]
    \centering
    \begin{minipage}[c]{0.70\textwidth}
        \centering
        \includegraphics[width=\linewidth]{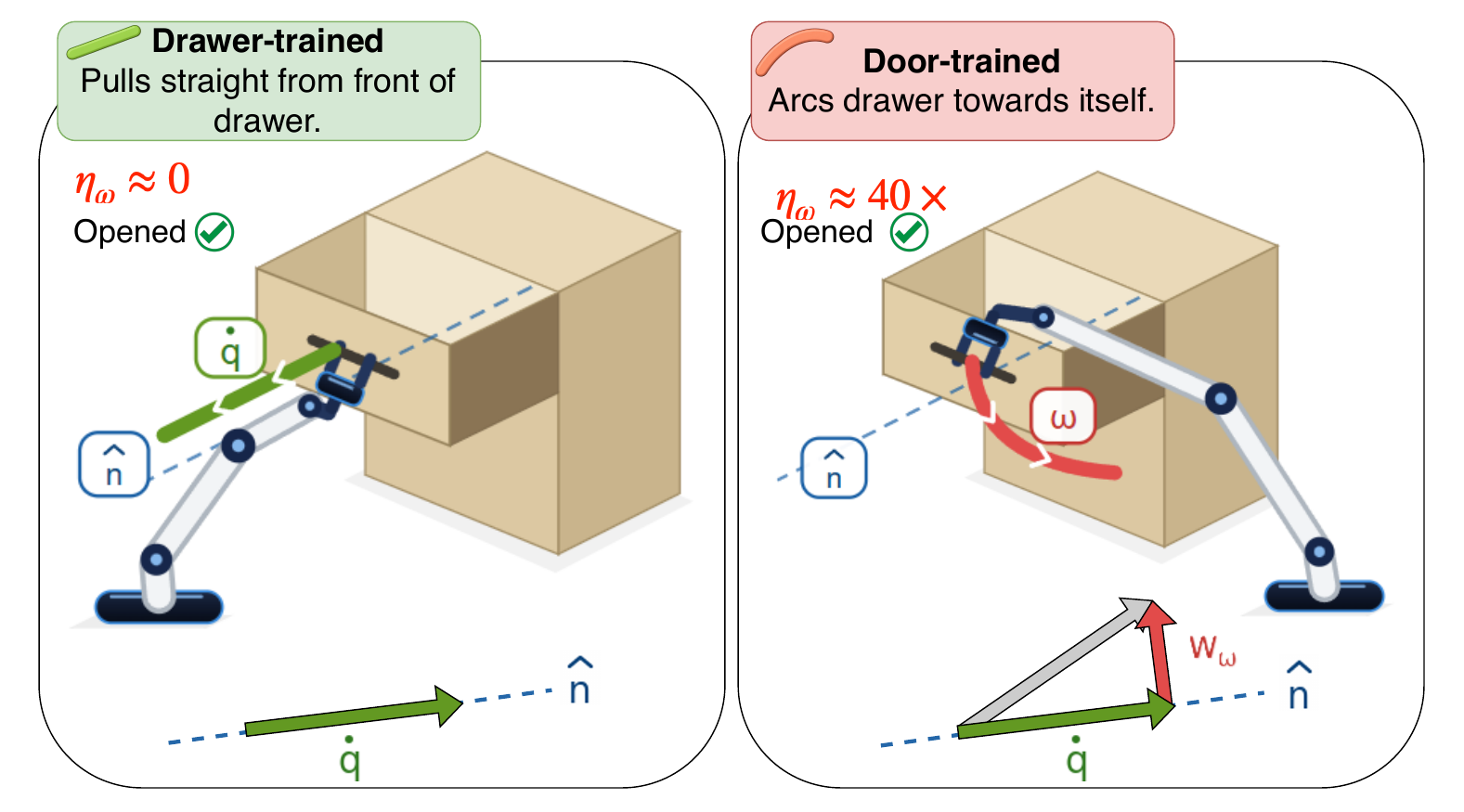}
    \end{minipage}
    \hfill
    \begin{minipage}[c]{0.26\textwidth}
        \caption{Success asks ``Did it open?''; CRS asks ``Was the commanded motion the one the mechanism admits?'' A drawer-trained and a door-trained policy both succeed at opening a held-out drawer. %Projected onto the drawer's joint axis $\hat{n}$, the drawer-trained policy's command is almost pure progress ($w_\omega \approx 0$), while the door-trained policy opens it only by exploiting slack, with $34\times$ the drawer-trained policy's off-axis waste.
        }
        \label{fig:CRS}
    \end{minipage}
\end{figure}

To provide this reading and complement \citep{xu2026metafine}, we introduce \emph{Constraint-Referenced Strategy} (CRS), which uses the kinematics of the task as the reference. CRS is based on the idea that, under a rigid grasp, the articulated  joint of an object defines exactly which end-effector motions advance it \citep{lynch2017modern}. %\audrey{ref applies only to the previous part not what follows?}, and
One screw axis %(i.e., the single rotation or translation axis that a joint permits)
encodes every valid opening mode. The reference is therefore mode invariant and needs no recorded
demonstration or learned action model. CRS scores a policy at reference states by projecting the commanded twist onto the joint's axis, yielding a signature with two channels: commanded progress along the joint and
off-constraint waste (Figure~\ref{fig:CRS}).
%\audrey{The next sentence is too long, with more than one idea. I would reformulate as ``''} %Across three transfer failures, success misses one and explains none; with verdicts that need only the joint model, and without demonstrations or closed-loop rollouts, CRS flags all three and tells them apart (Fig. \ref{fig:CRS} and Table \ref{tab:escalation}).
Our experiments show clear benefits of CRS against a binary success evaluation (Figure~\ref{fig:CRS} and Table~\ref{tab:escalation}).
% As a metric only earns findings after it earns trust, validation gates come first. Every verdict class is validated by a matched test against realized physics (ordering by $R^2$, direction by sign agreement above a measured noise floor, nulls by an equivalence band) and every analysis choice is fixed prior to evaluation, reported at worst case perturbations, with object level intervals throughout. \caro{to close out the introduction, we might add something like: only once these gates pass do we report what CRS reveals about transfer failures that binary success and existing diagnostics miss.}

% \begin{figure}[t]
% \centering
% \includegraphics[width=\linewidth]{figures/figure-CRS-v2.pdf}
% \caption{Illustration of the CRS method. Success asks "Did it open?", CRS asks: "Was the motion of the one the mechanism allows?" A drawer-trained and door-trained policy both succeed at opening a held-out drawer. Projecting each policy's commanded twist into the drawer's joint axis $\hat{\text{n}}$ shows that they are not equivalent. The drawer-trained policy's command is almost pure progress ($w_\omega \approxeq 0$), while the door-trained policy opens it only be exploiting slack, resulting in $44 \times$ the off-axis waste.}
% \label{fig:CRS}
% \end{figure}

\section{Constraint-Referenced Strategy (CRS)}
\label{sec:metric}

%\audrey{Be careful using the word ``metric'' as it has a special significance; validate that CRS respects what defines a metric (\url{https://en.wikipedia.org/wiki/Metric_space})}

CRS isolates the commanded gesture from confounds by evaluating a counterfactual: \emph{assuming the handle is already grasped at a valid pose, is the policy's commanded gesture kinematically consistent with the object's constraint?} We answer this through teacher-forced probing: a scripted controller, the oracle, places the robot at reference states along an opening sequence, and the policy's commanded end-effector twist is recorded at each. The reference states determine \emph{where} the policy is evaluated, while the joint geometry determines \emph{what} constitutes a valid command.

\paragraph{Decomposing the command.} Following standard rigid-body kinematics \citep{lynch2017modern}, a policy commands an end-effector  twist $\mathcal{V} = [\omega^\top, v^\top]^\top \in \mathbb{R}^6$, at the grasp point $c \in \mathbb{R}^3$: $\omega \in \mathbb{R}^3$ is the angular velocity % of the end-effector the same at every point of the rigid body, 
and $v \in \mathbb{R}^3$ is the linear velocity of $c$, all expressed in the robot base frame. The admissible motion is 1-DOF, with joint coordinate $q \in \mathbb{R}$ (an angle or a displacement)  oriented so that $\dot q > 0$ opens it. %, representing an angle for revolute joints ($q_\omega$) or a displacement for prismatic joints ($q_\text{pris}$). Its derivative $\dot{q} = dq/dt$ gives the instantaneous progress rate, with sign $s \in \{+1,-1\}$ chosen such that $s\dot{q}>0$ denotes motion toward the fully open limit.
The constraint geometry is specified by a unit vector $\hat{n}\in \mathbb{R}^3$ ($\|\hat{n}\| = 1$) along the joint axis
%\audrey{What is the format of a unit axis? Is this a canonical basis vector?}
and, for revolute joints, a point $o \in \mathbb{R}^3$ on that axis. %\audrey{Provide the space to which $o$ belongs.}.
%We call $\omega$ and $v$ the two \emph{action channels} of $\mathcal{V}$, rotation and translation, and read the joint progress from each. %\audrey{Action channel does not seem to be defined anywhere.}.
%\audrey{I suggest using paragraphs to help the reader separate how to proceed with revolute vs prismatic joints. It could be a good idea to merge back the definitions of waste in their respective paragraphs.}
For a revolute joint, the angular component directly measures wrist rotation about the joint axis: $\dot{q}_{\omega} = \omega^\top\hat{n}$ ($\mathrm{rad/s}$). 
% \begin{equation}
% \dot{q}_{\omega} = \omega^\top\hat{n}.
% \end{equation}
% The linear component provides an independent estimate of the same rate along the circular arc. Defining the lever arm $r = \|\hat{n}\times(c-o)\|$ and tangent $\hat{t} = (\hat{n}\times(c-o))/r$, we obtain:
% \begin{equation}
% \dot{q}_{v} = \frac{v^\top\hat{t}}{r}.
% \end{equation}
% Although both angular and linear components share the same units ($\mathrm{rad/s}$), we treat them independently. This enables CRS to score a policy as rotating about the wrong axis while its translation genuinely drags the joint along.%Both estimates are expressed in $\mathrm{rad/s}$, but we never average them: they read the same rate through different action channels, and keeping them apart is what allows a policy to be scored as rotating about the wrong axis while its translation genuinely drags the joint along.
For a prismatic joint, progress is the translation along the axis, $\dot{q}_{\mathrm{pris}}=v^\top\hat{n}$ in $\mathrm{m/s}$. %Thus, $\dot{q}_{v}\equiv\dot{q}_{\mathrm{pris}}$ and no rotational estimator is required.
The rotation the joint does not admit is \emph{waste}, $\eta_{\omega} = \|\omega - (\omega^\top\hat{n})\,\hat{n}\|$ ($\mathrm{rad/s}$), which reduces to $\eta_{\omega} = \|\omega\|$ for prismatic joints. These verdict-bearing channels depend only on the unit vector $\hat{n}$. For revolute joints, a second, translation-based estimate along the arc, $\dot{q}_{v} = v^\top\hat{t}/r$, with lever $r = \|\hat{n}\times(c-o)\| > 0$ and tangent $\hat{t} = \hat{n}\times(c-o)/r$, depends additionally on $o$ and $c$, and serves fidelity and diagnosis rather than verdicts.

\paragraph{Matched faithfulness gates.}
\label{sec:gates}

A reading of \emph{commanded} motion should be validated against its physical consequence. We therefore gate each progress verdict with a one-step counterfactual: from every reference state, we execute the command for one simulator step and compare the reading with the realized joint displacement $\Delta q_{\mathrm{real}}$. \emph{Ordering} claims are gated by the $R^2$ of a linear fit between $\Delta q_{\mathrm{real}}$ and $\dot q\,\Delta t$; \emph{direction} claims by sign agreement when $|\Delta q_{\mathrm{real}}|$ exceeds $\delta$, %\caro{Readers may confuse this with Kendall's $\tau$; might be nice to replace with $\delta$}
the 95th percentile under a null command; \emph{null} claims by an equivalence band $\pm\varepsilon$ of 5\% of the oracle rate. %Only the gates require a simulator; the metric itself needs only the command and the joint model. All scoring is teacher-forced at oracle states over $[\mathrm{onset},\mathrm{onset}+4]$; the window, channels, thresholds, and gate criteria are fixed before evaluation, and statistics are bootstrapped over objects ($10^4$ resamples, 95\% intervals).

\section{Experiments and Results}

\paragraph{Setup.}
\label{sec:setup}

We evaluate two policy families using their unmodified commands: OpenVLA-7B \citep{kim2024openvla} with rank-$32$ LoRA over three seeds, selecting the validation-minimum checkpoint, and a from-scratch point-cloud behavior-cloning (BC) policy over five seeds. %\audrey{Why different number of seeds? I would use 3 seeds each to uniformize.}
We test leave-one-mechanism-out generalization across three GAPartNet \citep{geng2023gapartnet} categories in ManiSkill \cite{taomaniskill3}:
\emph{door} (revolute, $39$ objects), \emph{drawer} (prismatic, $22$), and \emph{faucet} (revolute, $13$).
CRS is compared with binary success evaluation at $0.35$ opened fraction, $\Delta q_{\mathrm{real}}$, four command-side measures, DTW to demonstrations, and MetaFine's stability (Table~\ref{tab:escalation}).

%\audrey{Table 1 is difficult to understand. What are the numbers? I can't follow the description of null, catch etc in the table caption. I suggest having a little text paragraph here saying ``For each evaluation strategy considered, we report the evaluation score along with its interpretation. Specifically, we label as: \textit{catch} when ...; \textit{mislead} when; \textit{miss} when...; or \textit{null} when...''}

\paragraph{Instrument validation.} %\audrey{Unclear how to relate the following to Table 1. Is this a discussion of Table 1's results? Or is it something that we should read before Table 1 to help understand it? Move the paragraph before the description of Table 1 is that's the case}
The commanded rate tracks realized displacement ($R^2 = 0.69/0.89/0.65$ for door/drawer/faucet), and command and realized directions agree at 1.000 above the noise floor on oracle commands. Calibration slopes remain below one (0.49/0.28/0.60), so rates are read as
directional and relative. As a null control (silent stall), the drawer-trained policy commands no revolute motion on doors: $|\dot q_\omega|$ is $12.9\times$ smaller than $\varepsilon$.

\paragraph{Detection labels.}
Table~\ref{tab:escalation} reads each measure against the scripted oracle on the target mechanism, with the failure direction fixed per measure. \textbf{catch}: beyond the oracle's 95\% CI in the failure direction;
\emph{misleads}: within the CI despite a wrong gesture; miss: beyond the CI in the other direction. The CRS signature pairs the sign of signed progress (null within $\pm\varepsilon$) with whether waste is flagged.

\begin{table}[t]
\centering
\scriptsize
\setlength{\tabcolsep}{3.5pt}
\setlength{\aboverulesep}{0pt}
\setlength{\belowrulesep}{0pt}
\setlength{\extrarowheight}{1.5pt}

\caption{Which measure detects each transfer failure, and which attributes
it. Labels: Sec.~3; silent-stall is a null control; in the false-pass
column, success reads 1 on 59 of 110 episodes. Per-object means over
retained reference states, averaged over objects; 95\% CIs bootstrap over
objects. OpenVLA: seed 0 (reverse-push signs replicate on seeds 1 and 2),
teacher-forced except success. BC: 5 seeds pooled, closed loop (CL).
Signed progress and $|\dot q|$ in rad/s (door, faucet) or m/s (drawer);
$\Delta q_{\mathrm{real}}$ in rad or m per step; $\|\omega\|$, $\eta_\omega$
in rad/s. --: not computed. Advancing: fraction of states advancing the joint; alignment: $|\omega^\top\hat n|/\|\omega\|$ (revolute) or $|v^\top\hat n|/\|v\|$ (prismatic).}
\label{tab:escalation}
\begin{adjustbox}{max width=\linewidth}
\begin{tabular}{llcccc}
\toprule
family & measure & silent-stall & reverse-push & misaimed & false-pass \\
 & & drawer$\to$door & door$\to$drawer & door$\to$faucet & BC door$\to$drawer \\
\midrule
\multirow{2}{*}{outcome}
 & success@0.35 (CL) & 0.000 & 0.091 \textbf{catch} & 0.026 \textbf{catch} & 0.536 \textbf{catch} \\
 & $\Delta q_{\mathrm{real}}$ (one step) & 0.0075 & $-0.0003$ \textbf{catch} & 0.027 \emph{misleads} & 0.0045 \emph{misleads} \\
\midrule
\multirow{4}{*}{\shortstack[l]{command\\scalar}}
 & rotation $\|\omega\|$ & 0.023 & 0.031 \emph{misleads} & 1.00 \emph{misleads} & 1.29 \textbf{catch} \\
 & unsigned rate $|\dot q|$ & 0.014 & 0.270 \emph{misleads} & 0.711 \textbf{catch} & 0.360 \emph{misleads} \\
 & advancing & 0.47 & 0.005 \textbf{catch} & 0.42 \textbf{catch} & 1.00 \emph{misleads} \\
 & alignment & 0.68 & 0.984 \textbf{catch} & 0.57 \textbf{catch} & 0.963 \textbf{catch} \\
\midrule
demonstration & DTW to demos & -- & -- & -- & 0.31 \textbf{catch} \\
\midrule
execution & MetaFine stability~\citep{xu2026metafine} & 0.93 & 0.95 miss & 0.88 \emph{misleads} & 0.89 \textbf{catch} \\
\midrule
\rowcolor{crsblue}
 & signed $\dot q_{\omega}$ / $\dot q_{\mathrm{pris}}$ & $-0.0014$ & $-0.264$ \textbf{catch} & $-0.14$ \textbf{catch} & $+0.36$ \emph{misleads} \\
\rowcolor{crsblue}
 & waste $\eta_{\omega}$ & 0.016 & 0.031 \emph{misleads} & 0.38 \textbf{catch} & 1.29 \textbf{catch} \\
\rowcolor{crsblue}
\multirow{-3}{*}{\textbf{CRS (ours)}}
 & \textbf{signature (sign, $\eta_{\omega}$)} & \textbf{null} & \textbf{reversed} & \textbf{off-axis ($-$)} & \textbf{off-axis ($+$)} \\
\bottomrule
\end{tabular}
\end{adjustbox}
\end{table}

\paragraph{A pass on the wrong gesture.}
A from-scratch BC policy trained on doors and rolled out on drawers (false-pass column) succeeds in 59 of 110 episodes, on 20 of 22 drawers. Success registers a drop relative to a drawer-trained policy of the same
architecture (0.54 vs.\ 0.77) but reads it as degraded competence. In the episodes it completes, the door-trained policy commands $1.27$~rad/s
$[1.19, 1.34]$ of off-axis rotation (per-episode means), against $0.032$~rad/s $[0.028, 0.037]$ for the drawer-trained policy in its own successful episodes: about $40\times$ at matched outcome. The success bit
reads 1 on a gesture the joint does not admit. This waste does not differ between successful and failed episodes ($1.27$ vs.\ $1.26$~rad/s): it is a
property of the gesture, not of the outcome. On a prismatic joint, waste equals $\|\omega\|$, so $\|\omega\|$ and DTW also flag this case;
$\|\omega\|$ misleads on revolute joints (misaimed).
%The drawer tolerates this extra rotation, so the outcome remains successful even though the gesture is wrong. Here the waste signal is especially direct: for a prismatic joint, $\omega_\parallel=0$ and therefore $w_\omega=|\omega|$, independent of the estimated axis, origin, or contact. Two further transfer failures, invisible respectively to command-side scalars and to progress measures, are documented in \ref{app:cases}.

\paragraph{Reversed and off-axis commands.}
Trained on doors and probed on held-out drawers (reverse-push column), the OpenVLA policy commands the drawer shut: the commanded rate is negative on
all three seeds, yet $\|\omega\|$ matches the drawer oracle and alignment, being unsigned, departs only slightly (0.984 vs.\ 0.999). The drawer does
not move within one step, so $\Delta q_{\mathrm{real}}$ shows a stall, not a reversal; the reversal stays command-side. The same door$\to$drawer
shift thus produces two different failures: a commanded closing with OpenVLA and a false-pass with BC. Success reads a lower rate for both and explains neither. On faucets (misaimed column), the same policy reaches
68\% of the oracle's realized displacement, so $\Delta q_{\mathrm{real}}$ reads healthy. Yet it rotates about an axis the faucet lacks, with waste $5.4\times$ the oracle's. Its commanded angular rate is nonetheless
negative and fails the direction gate, consistent with the linear component driving the joint through a grasp that is not fully rigid. Closed-loop success flags this case (0.026) but cannot say why. Several measures flag each failure, and alignment flags all three. None attributes them: alignment is invariant to the sign of the command, so a drawer pushed shut reads like one pulled open (0.984 vs.\ 0.963), and success cannot tell a reversed command from a slow or blocked one. Signed progress separates reversed from advancing commands, and waste separates on-axis from off-axis ones; together they place the three failures in distinct cells (Table~\ref{tab:escalation}, last row).

\paragraph{Robustness to estimated kinematics.}
Since real robots lack an exact joint model, we re-score all rollouts with corrupted kinematics while keeping the commanded twists and realized physics fixed: axis tilts up to $30^\circ$ and axis-point and contact
offsets up to 5~cm, worst case over 20 directions. The verdict-bearing channels remain stable through the full $30^\circ$ perturbation: the three
failures and the null control keep the same verdict. The lever-based channel $\dot q_v$ is less robust (door $R^2$: $0.90 \to 0.34$). The misaimed verdict holds relative to the oracle scored under the same axis;
as an absolute threshold it tolerates about $8^\circ$.

% \paragraph{What changes closed-loop.}
% On its own states, the door-trained policy instead advances the drawer while still commanding door-scale angular waste. Among the $8/22$ instances \caro{with 8/22 drawers reaching a scoreable window, is that subset random or biased toward a particular failure mode? If so, the $\dot q_{\mathrm{pris}}=+0.31$ may be describing how the door-trained policy behaves on a subset of drawers lenient enough to move at all under a bad gesture. This might be worth commenting on to avoid reviewer complaints.} reaching a scoreable window, $\dot q_{\mathrm{pris}}=+0.31$ and $w_\omega=1.53$. The failure therefore changes from wrong direction to wrong gesture, which is why the Case 1 verdict is reported as teacher-forced.

% The grasp mask separates two closed-loop failures that success alone cannot distinguish. The drawer-trained policy on doors fails to establish a grasp ($\mathrm{grasp\_frac}=0.13$) \caro{changed from $\mathrm{grasp_frac}$ to $\mathrm{grasp\_frac}$ to match above}, while the door-trained policy on drawers retains its grasp ($0.46$) but fails by gesture. Both have near-zero success, but for mechanically different reasons. %The full teacher-forced-to-closed-loop grid is given in \ref{app:closedloop}.

\section{Conclusion and Discussion}
CRS reads a policy's commanded gesture against the joint it acts on. It attributes failures that success registers but cannot explain, and exposes
passes obtained with a gesture the joint does not admit. Because CRS scoring requires only the joint axis, it could complement per-stage success and trajectory stability in MetaFine's behavior dimension~\citep{xu2026metafine}, given annotated joint axes. Limitations: CRS assumes a rigid grasp, which the faucet case suggests is not always met, and a single 1-DOF joint. It scores angular waste only, is validated
in simulation on three mechanisms, and treats waste as a command-side quantity whose physical cost we have not measured. Future work includes other lower pairs (helical, cylindrical, planar) and real hardware, where
the gate could read realized progress from proprioception under a rigid grasp, the axis could be estimated from RGB-D video, and relative verdicts
would use an in-domain policy scored with the same axis as reference.%Two extensions are natural: other lower pairs (helical, cylindrical, planar), and real hardware, where the gate can read realized progress from proprioception under a rigid grasp and the axis can be estimated from RGB-D video, which the $30^{\circ}$ margin makes plausible for relative verdicts.

\bibliography{example}

@inproceedings{li2024simpler, author={Li, X. and Hsu, K. and Gu, J. and others}, title={Evaluating Real-World Robot Manipulation Policies in Simulation}, booktitle={CoRL}, year={2024}}

@inproceedings{li2023dtw, author={Li, C. and Vlastelica, M. and Blaes, S. and others}, title={Learning Agile Skills via Adversarial Imitation of Rough Partial Demonstrations}, booktitle={CoRL}, year={2022}}

@article{elmallah2025scorestepsnotjust, author={ElMallah, R. and Chhajer, K. and Lee, C.-G.}, title={Score the Steps, Not Just the Goal: {VLM}-Based Subgoal Evaluation for Robotic Manipulation}, journal={arXiv:2509.19524}, year={2025}}

@inproceedings{duan2025aha, author={Duan, J. and Pumacay, W. and Kumar, N. and others}, title={{AHA}: A Vision-Language-Model for Detecting and Reasoning over Failures in Robotic Manipulation}, booktitle={ICLR}, year={2025}}

@article{xu2026metafine, author={Xu, H.-Y. and Zhang, P. and Ge, Z. and others}, title={Beyond Binary Success: A Diagnostic Meta-Evaluation Framework for Fine-Grained Manipulation}, journal={arXiv:2605.19986}, year={2026}}

@book{lynch2017modern, author={Lynch, K. M. and Park, F. C.}, title={Modern Robotics: Mechanics, Planning, and Control}, publisher={Cambridge University Press}, year={2017}}

@inproceedings{kim2024openvla, author={Kim, M. J. and Pertsch, K. and Karamcheti, S. and others}, title={{OpenVLA}: An Open-Source Vision-Language-Action Model}, booktitle={CoRL}, year={2024}}

@inproceedings{geng2023gapartnet, author={Geng, H. and Xu, H. and Zhao, C. and others}, title={{GAPartNet}: Cross-Category Domain-Generalizable Object Perception and Manipulation via Generalizable and Actionable Parts}, booktitle={CVPR}, year={2023}}

@article{taomaniskill3,
  title={ManiSkill3: GPU Parallelized Robotics Simulation and Rendering for Generalizable Embodied AI},
  author={Stone Tao and Fanbo Xiang and Arth Shukla and others},
  journal = {RSS},
  year={2025},
}
\clearpage

\end{document}